\documentclass[runningheads]{llncs}
\usepackage{graphicx}
\usepackage{multirow}
\usepackage{booktabs}
\usepackage{tabularx}
\usepackage{amsmath}
\usepackage{amsfonts}
\usepackage{xspace}
\usepackage{color}
\usepackage{bm}
\usepackage{marvosym}

\begin{document}

\bibliographystyle{splncs04}

%



\title{CountMamba: Exploring Multi-directional Selective State-Space Models for Plant Counting}
\titlerunning{CountMamba}
%
\author{Hulingxiao He\inst{1} \and 
Yaqi Zhang\inst{2} \and 
Jinglin Xu\inst{2} \and 
Yuxin Peng (\Letter) \inst{1}}
%
\authorrunning{Hulingxiao He et al.}
%

\institute{Peking University, Beijing 100091, China \\
\email{hehulingxiao@stu.pku.edu.cn, pengyuxin@pku.edu.cn} \and
University of Science and Technology Beijing, Beijing 100083, China\\
\email{M202320870@xs.ustb.edu.cn, xujinglinlove@gmail.com} }

\maketitle              
\begin{abstract}

Plant counting is essential in every stage of agriculture, including seed breeding, germination, cultivation, fertilization, pollination yield estimation, and harvesting. Inspired by the fact that humans count objects in high-resolution images by sequential scanning, we explore the potential of handling plant counting tasks via state space models (SSMs) for generating counting results. In this paper, we propose a new counting approach named CountMamba that constructs multiple counting experts to scan from various directions simultaneously. Specifically, we design a Multi-directional State-Space Group to process the image patch sequences in multiple orders and aim to simulate different counting experts. We also design Global-Local Adaptive Fusion to adaptively aggregate global features extracted from multiple directions and local features extracted from the CNN branch in a sample-wise manner. Extensive experiments demonstrate that the proposed CountMamba performs competitively on various plant counting tasks, including maize tassels, wheat ears, and sorghum head counting.

\end{abstract}
\section{Introduction}
Plant counting is indispensable at nearly every critical stage of agricultural production, spanning from seed breeding \cite{guo2018aerial}, germination \cite{primicerio2017individual}, cultivation \cite{lu2017tasselnet,xiong2019tasselnetv2,lu2021tasselnetv3}, fertilization \cite{boissard2008cognitive}, pollination yield estimation \cite{bai2023rice}, and harvesting \cite{jin2019high}. Moreover, it plays a vital role in extracting plant characteristic growth and typical plant features such as the number of leaves \cite{giuffrida2016learning}, corn ears \cite{lu2017tasselnet}, and wheat spike \cite{xiong2019tasselnetv2} which can be utilized for yield prediction \cite{bai2023rice}, seed quality testing \cite{donapati2023real}, and verification of the impact of novel genes \cite{wang2022improving}. 

However, traditional yield estimates are predominantly predicted manually, a process characterized by its time-consuming and labor-intensive nature,  influenced by numerous subjective factors. Given these drawbacks, there is an urgent need for automatic plant counting to reduce human error, and provide more accurate and timely insights into agricultural productivity.

Agricultural practitioners have tried to automate plant counting tasks over the past decade. Previously, distinguishing plants relied on color contrast \cite{lu2016region}, but it became challenging with obstructions or similar colors. Recently, deep learning-based methods eased this with segmentation \cite{donapati2023real} or detection \cite{liu2022yolov5} pipelines through Faster R-CNN \cite{ren2015faster} and fully convolutional networks (FCN) \cite{long2015fully}. Although existing methods boost the performance of plant counting, they still encounter two challenges: \textbf{1) Arbitrary Direction:} Since the images for plant counting are usually captured from a top-down perspective, the plants can be distributed in an arbitrary direction, like in the horizontal, vertical, diagonal or anti-diagonal direction. Extracting spatial features along the direction of plant distribution could facilitate counting. \textbf{2) High Resolution:} Since high-resolution imagery is frequently used in modern plant phenotyping platforms such as UAVs (unmanned aerial vehicles), it is inefficient to model a long-term dependency for image patches.

Following the breakthrough of Mamba \cite{gu2023mamba}, a state-of-the-art model integrating selective scanning (S6), there has been a notable surge in leveraging State Space Models (SSMs) across diverse computer vision tasks \cite{huang2024localmamba,zhao2024rs,zhu2024vision}. Inspired by the fact that humans  count objects in high-resolution images by doing sequential scan following the direction of the plant distribution, Mamba serves as a potential resolution for counting plants efficiently and effectively. 

To explore the potential of plant counting tasks via SSMs, we propose CountMamba, a simple but effective model, to adapt Mamba for plant counting. CountMamba is formulated with three principal components. Specifically, the \textbf{1) Multi-directional State-Space Group} (MSSG) performs with four parallel branches to extract features from various directions, with each branch consisting of stacked Horizontal State-Space Blocks (HSSBs), Vertical State-Space Blocks (VSSBs), Diagonal State-Space Blocks (DSSBs) and Anti-diagonal State-Space Blocks (ASSBs), respectively. Then, the \textbf{2) Global-local Adaptive Fusion} aggregates the global features from MSSG adaptively in a sample-wise manner and employs an CNN branch to complement the global features with local information. Finally, the \textbf{3) Counter and Normalizer} are utilized to infer the number of plants based on the normalized count map. Through adaptively integrating local relationships and global information along appropriate directions, our CountMamba achieves competitive counting results on several plant counting tasks, including maize tassels counting, wheat ears counting, and sorghum head counting.


Our contributions can be summarized as follows:
\vspace{-3mm}
\begin{itemize}
\item We propose CountMamba, a new model introducing the state space models to perform plant counting tasks effectively and efficiently.
\item We propose the Multi-directional State-Space Group that extracts features of patch sequences in various orders and boosts Mamba to adapt to plants distributed in any direction.
\item We propose the Global-Local Adaptive Fusion that adaptively aggregates local information and global information from appropriate directions to select informative features in a sample-wise manner.
\item Extensive experiments on various plant counting benchmarks demonstrate that our CountMamba can provide a powerful and promising backbone for plant counting.
\end{itemize}

\section{Related Work}
\subsection{Plant Counting}
Counting the number of plants significantly predicts crop yield and other aspects. Initially, the counting of plants is usually distinguished from other backgrounds through different colors \cite{lu2016region} in the agricultural field, as there exists a strong color contrast between different plants.
Nevertheless, this method is difficult to distinguish when plants are obstructed or have similar colors. Based on this issue, many methods attempt to solve the problem through phenotypic parameters such as plant appearance texture features \cite{cointault2008field}. However, distinguishing phenotypic parameters was also difficult at the time. With the emergence of deep learning methods, the difficulty is mainly alleviated.

In deep learning, many methods tend to construct segmentation \cite{donapati2023real} or detection \cite{liu2022yolov5,madec2019ear,zhao2023small} pipelines to locate plants. Because both detection and segmentation have well-known, powerful, and easy-to-use frameworks, such as faster R-CNN \cite{ren2015faster} and fully convolutional networks (FCN) \cite{long2015fully}, apart from the traditional convolutional networks, new network architectures such as Transformer \cite{vaswani2017attention} and Mamba \cite{gu2023mamba} that have emerged in recent years can also be utilized for extracting plant phenotypic characteristics. Harada et al. \cite{sho2023hybrid} exploit a novel hybrid wheat detection model by incorporating the CNN and Transformer for modeling long-range dependence. WheatNet \cite{zhao2023small} consists of two parts: a Transform Network, which reduces the effect of differences in the color features, and a Detection Network, which is designed to improve the capability of detection. FlowerNet \cite{lin2024framework} is a framework for flower counting based on the algorithm of YOLACT++ \cite{zhou2020yolact++}, a real-time instance segmentation model. TasselNet \cite{lu2017tasselnet} is the first work of a local count network applied to plants. TasselNetV2 \cite{xiong2019tasselnetv2} discovers weak context on the plants, which is essential for the counting work.

Apart from the segmentation and detection abundantly utilized in plant counting, regression-based plant counting is less adopted. Giuffrida et al. \cite{giuffrida2016learning} introduce scale and rotation invariance to learn features in a log-polar representation, which aims to count leaves in round shapes. A combined network that integrates density map regression and image segmentation is proposed by Wu et al. \cite{wu2019automatic} for rice seedlings counting. Regression-based methods generally use the overall features of the image for regression analysis, ignoring the spatial information of the image. Therefore, their performance is worse than that of some density map-based methods constructed by detection or segmentation pipelines.

\subsection{State Space Models}

State Space Models (SSMs) have shown remarkable and efficient ability in capturing long sequences by utilizing selective state space to capture relevant information. Unlike Transformer \cite{vaswani2017attention}, SSMs have lower linear time complexity and implement linear time running in terms of sequence length, making them particularly suitable for handling very long sequences. Gu et al. \cite{gu2021efficiently} first introduce a Structured State-Space Sequence model (S4) engineered with a specific focus on capturing long-range dependencies, which boasts the advantage of linear complexity. Inspired by this, various models have been developed, such as the S5 layer \cite{smith2022simplified}, which introduces MIMO SSM and efficient parallel scan into the S4 layer. Fu et al.  \cite{fu2022hungry} propose a novel SSM layer, H3, which significantly narrows the performance disparity between SSMs and Transformer attention in language modeling. Methea et al. \cite{mehta2022long} enhance the expressiveness of the Gated State Space layer on S4 by incorporating additional gating units, thereby boosting its performance. Lately, Gu et al. \cite{gu2023mamba} have introduced a data-dependent SSM layer and developed Mamba, a universal language model backbone. Mamba exhibits superior performance compared to Transformers across various sizes when trained on large-scale real-world data while demonstrating linear scalability with sequence length. Based on this work, Huang et al. \cite{huang2024localmamba} solve the simple local 2D dependency relationship in the original Mamba model by dividing the image into different windows, effectively capturing local dependency relationships while maintaining Mamba's \cite{gu2023mamba} original global dependency ability. RS-Mamba \cite{zhao2024rs} is proposed to process high-resolution remote sensing images. Vision-Mamba \cite{zhu2024vision} combines bidirectional SSM for global visual context modeling for data dependency and position embedding for position-aware visual recognition.

\section{Methodology}

\subsection{Preliminary}
State Space Models (SSMs) and various models based on Mamba are inspired by the continuous system, which maps the 1-D function or sequences $x(t)\!\in\!\mathbb{R}\rightarrow\!y(t)$ through a hidden state $h(t)$ $\in$ $\mathbb{R}^{N}$. Formally, SSMs employ the following ordinary differential equation (ODE) to model the input data:
\begin{equation}\label{ori1}
\begin{aligned}
& h'(t) = \textbf{A}h(t) +\textbf{B}x(t) \\
& y(t) = \textbf{C}h(t)
\end{aligned}
\end{equation}
where $\textbf{A}\!\in\!\mathbb{R}^{N\times N}$ represents the evolution matrix, $\textbf{B}\!\in\!\mathbb{R}^{N\times 1}$ and $\textbf{C}\!\in\!\mathbb{R}^{N\times 1}$ indicate the projection matrices.
SSMs approximate this continuous ODE through discretization techniques. The S4 and Mamba are the discrete versions of the continuous system, which use the timescale parameter $\bm{\Delta}$ to convert continuous parameters $\mathbf{A}$ and $\mathbf{B}$ into discrete parameters $\overline{\mathbf{A}}$ and $\overline{\mathbf{B}}$. The commonly used method for transformation is zero-order hold (ZOH), which is defined as follows:
\begin{equation}
\begin{aligned}
& \overline{\mathbf{A}}=\exp(\bm{\Delta} \mathbf{A}) \\
& \overline{\mathbf{B}}=(\bm{\Delta}\mathbf{A})^{-1}(\exp (\bm{\Delta}\mathbf{A})-\mathbf{I})\cdot\bm{\Delta}\mathbf{B}. \\
\end{aligned}
\end{equation}
After the discretization of $\overline{\mathbf{A}}$ and $\overline{\mathbf{B}}$, the discretized version of Equation (\ref{ori1}) using a step size $\bm{\Delta}$ can be rewritten as:

\begin{equation}\label{ori2}
\begin{aligned}
    &h_{t}=\overline{\mathbf{A}} h_{t-1}+\overline{\mathbf{B}} x_{t} \\
    &y_{t}=\mathbf{C} h_{t}. \\
\end{aligned}
\end{equation}
To enhance computational efficiency, the iterative process described in Equation (\ref{ori2}) can be accelerated using parallel computation techniques, leveraging a global convolution operation:
\begin{equation}
\begin{aligned}
\boldsymbol{y}=\boldsymbol{x}\otimes\overline{\bm{K}},\quad \overline{\bm{K}}=(\mathbf{C} \overline{\mathbf{B}}, \mathbf{C}\overline{\mathbf{A}}\overline{\mathbf{B}}, \ldots, \mathbf{C}\overline{\mathbf{A}}^{L-1}\overline{\mathbf{B}})
\end{aligned}
\end{equation}
where $\otimes$ is the convolution operation, $L$ is the length of the input sequence $\bm{x}$, and $\overline{\boldsymbol{K}}\!\in\!\mathbb{R}^{L}$ is the kernel of the SSM.

\begin{figure}[t]
    \centering
    \includegraphics[width=0.99\linewidth]{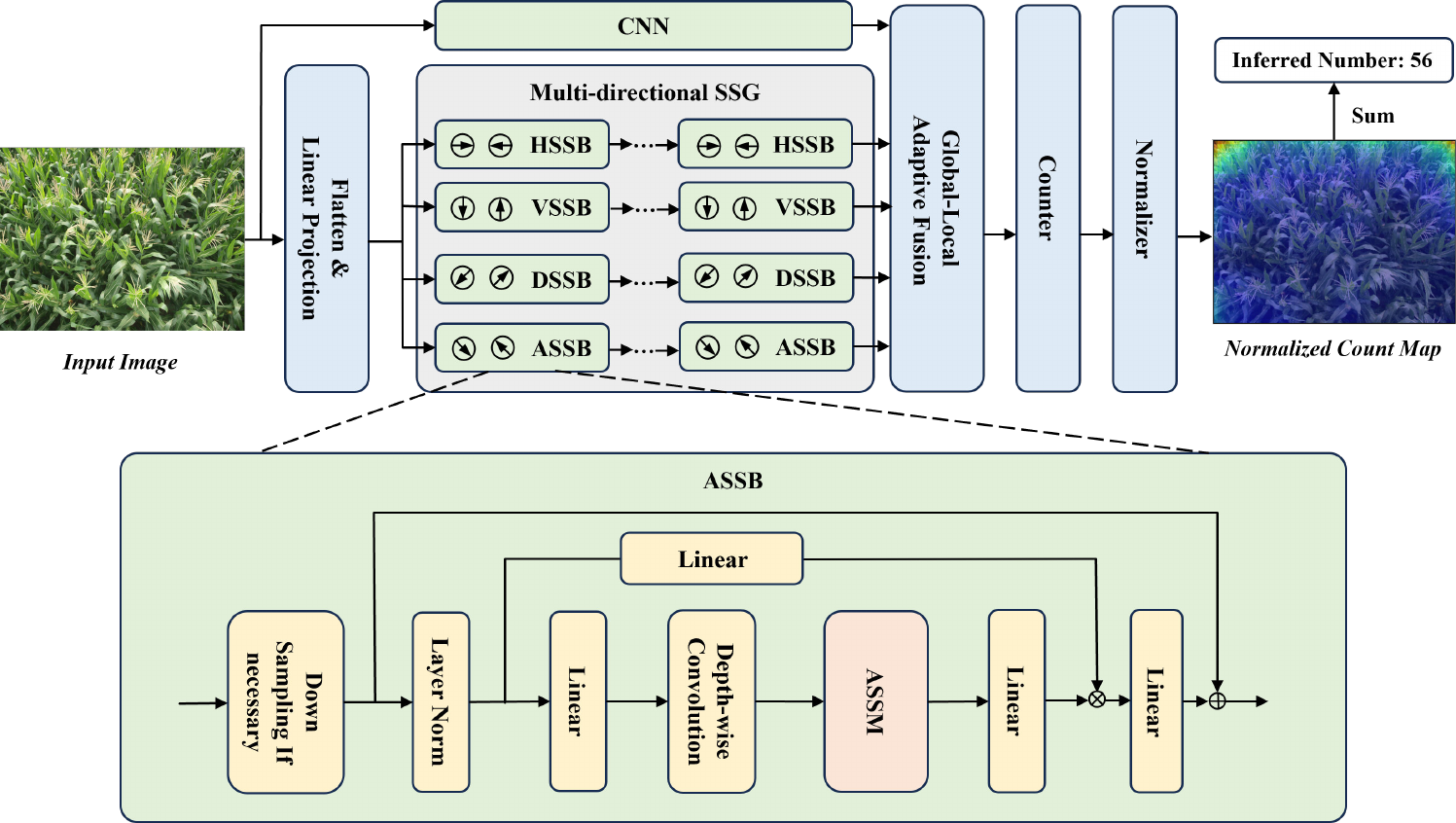}
    \caption{An overview of the proposed CountMamba. It contains a Multi-directional State-Space Group (MSSG) comprised of stacked Horizontal State-Space Blocks (HSSBs), Vertical State-Space Blocks (VSSBs), Diagonal State-Space Blocks (DSSBs), and Anti-diagonal State-Space Blocks (ASSBs) in parallel, followed by Global-Local Adaptive Fusion, Counter and Normalizer to achieve plant counting.}
    \label{fig:pipeline}
\end{figure}

\subsection{Overall Architecture}
As shown in Fig. \ref{fig:pipeline}, CountMamba consists of three main components: \textbf{Multi-directional State-Space Group (MSSG)}, \textbf{Global-Local Adaptive Fusion (GLAF)}, \textbf{Counter}, and \textbf{Normalizer}. Given an input image $\mathbf{I}\!\in\!\mathbb{R}^{H\times W\times 3}$, we flatten it into a patch sequence and then project it to a feature sequence via linear transformation. Subsequently, we employ a Multi-directional State-Space Group (MSSG) to acquire the down-sampled deep features $\bm{F}_H$, $\bm{F}_V$, $\bm{F}_D$, and $\bm{F}_A$ from stacked Horizontal State-Space Blocks (HSSBs), Vertical State-Space Blocks (VSSBs), Diagonal State-Space Blocks (DSSBs) and Anti-diagonal State-Space Blocks (ASSBs). $\bm{F}_H$, $\bm{F}_V$, $\bm{F}_D$, and $\bm{F}_A$ are then fused adaptively in a sample-wise manner and refined by local features extracted by the CNN branch. Finally, the aggregated features undergo the Counter and Normalizer to predict the normalized counting map $\mathbf{C}_{n}\!\in\!\mathbb{R}^{H\times W}$, which is summed up to infer the overall number of plants $C$ in the image.

\subsection{Multi-directional State-Space Group}
Multi-directional State-Space Group (MSSG) is designed to extract plant image features from various directions to adapt to different distributions. As shown in Fig. \ref{fig:pipeline}, MSSG is comprised of four parallel branches, consisting of stacked Horizontal State-Space Blocks (HSSBs), Vertical State-Space Blocks (VSSBs), Diagonal State-Space Blocks (DSSBs), and Anti-diagonal State-Space Blocks (ASSBs). HSSB, VSSB, DSSB, and ASSB share the same architecture except for the scan direction, possessing a Horizontal State-Space Module (HSSM), Vertical State-Space Module (VSSM), Diagonal State-Space Module (DSSM), Anti-diagonal State-Space Module (ASSM), respectively. The details of HSSM, VSSM, DSSM, and ASSM are illustrated in Fig. \ref{fig:components}.

Inspired by the design of \cite{zhao2024rs}, a layer normalization is adopted to standardize the input data for balancing computational efficiency and capability, followed by a linear layer for transformation. Then, the depth-wise convolution operates on each channel separately to extract local features. The features are fed into State-Space Modules with corresponding directions to perform selective scanning in both the forward and backward directions, simulating an expert counting in a preferred scanning order. In addition, the output is linearly transformed and undergoes a gating operation with outputs of a linear transformation of the normalized features. The features undergo another linear layer and are added with inputs through a residual connection.

\subsection{Global-Local Adaptive Fusion}
\begin{figure}[t]
    \centering
    \includegraphics[width=0.99\linewidth]{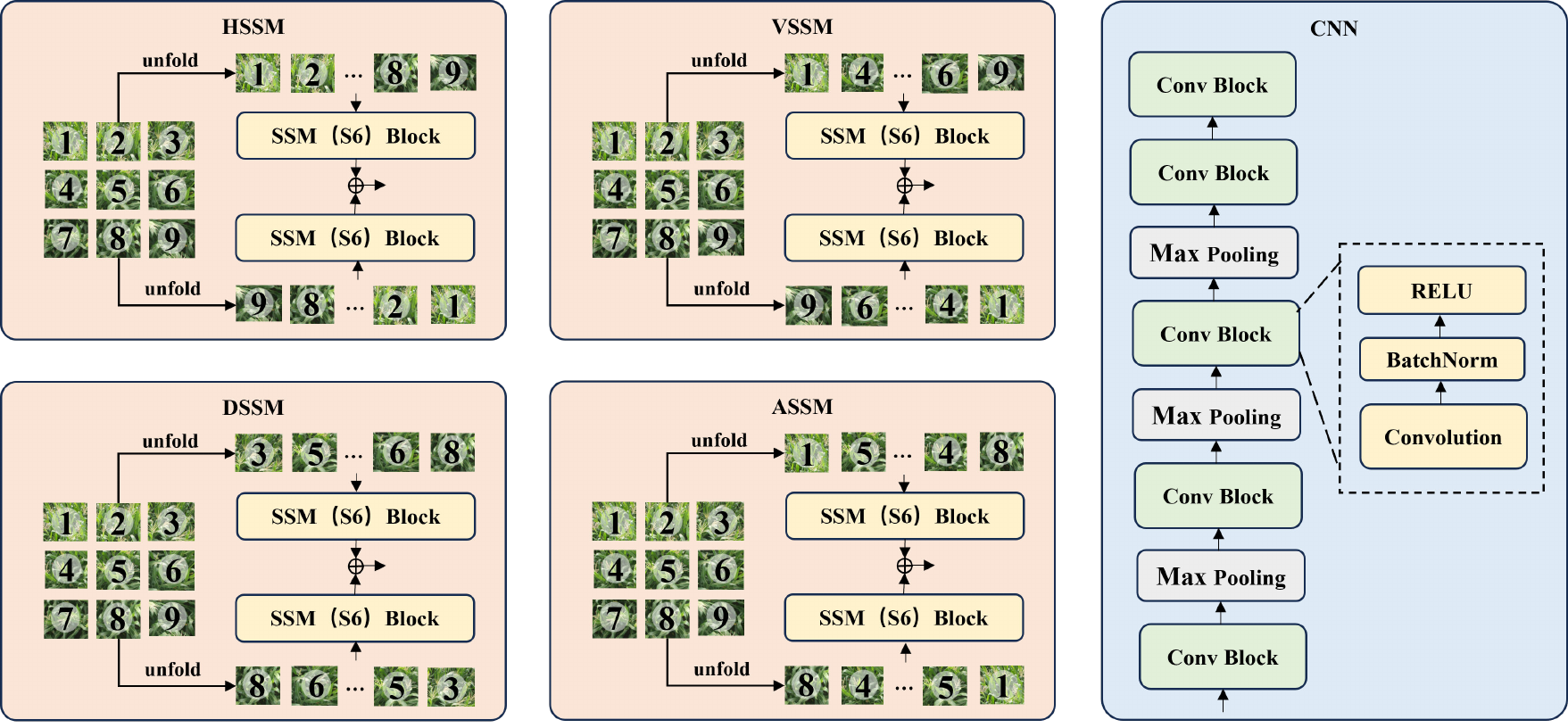}
    \caption{Illustration of the structure of HSSM, VSSM, DSSM, ASSM, and CNN branch.}
    \label{fig:components}
\end{figure}

As plants in different images can be distributed in any direction, we design a sample-wise fusion mechanism to adaptively aggregate features from different directions, which can be defined as:
\begin{equation}
    \bm{F}_{\text{global}}=\alpha_{H}\circ\bm{F}_H + \alpha_{V}\circ\bm{F}_V + \alpha_{D}\circ\bm{F}_D + \alpha_{A}\circ\bm{F}_A
\end{equation}
where $\bm{F}_{\text{global}}$ represents the adaptively aggregated feature embedding, the element-wise multiplication denoted by $\circ$, and $\bm{F}_H, \bm{F}_V, \bm{F}_D, \bm{F}_A$ refer to the extracted featured from HSSM, VSSM, DSSM, and ASSM, respectively. The adaptive fusion weights $\alpha_{H}$, $\alpha_{V}$, $\alpha_{D}$, and $\alpha_{A}$ are defined as:
\begin{equation}
    \alpha_{H}, \alpha_{V}, \alpha_{D}, \alpha_{A} = \text{softmax}(\mathbf{W}\cdot\text{Concat}(\bm{F}_H,\bm{F}_V,\bm{F}_D,\bm{F}_A))
\end{equation}
where $\mathbf{W}$ is a learnable linear transformation. Moreover, we add a lightweight CNN branch that contains stacked layers of convolution, batch normalization, non-linear activation, and max pooling to extract local features $\bm{F}_{\text{local}}$. The detailed architecture of the CNN branch is illustrated in Fig. \ref{fig:components}. Finally, the aggregated features from multiple directions $\bm{F}_{\text{global}}$ are refined by $\bm{F}_{\text{local}}$ to obtain the final features $\bm{F}_{\text{fused}}$, defined as:
\begin{equation}
    \bm{F}_{\text{fused}} = \bm{F}_{\text{global}} + \beta \cdot \bm{F}_{\text{local}}
\end{equation}
where $\beta$ is the hyper-parameter to control the weight of local information.

\subsection{Counter and Normalizer}
Counters are used to count plants in a specified area, while normalizers address the impact of duplicate counting of two overlapping images during the counting process \cite{xiong2019tasselnetv2}. After extracting the features of the image $\bm{F}_{\text{fused}}$, it needs to be sent to the counter for quantity prediction. In our experiment of the regression, a count map $C_r$ is used for the image transformation following \cite{xiong2019tasselnetv2}, with each count representing a local region sized $r\!\times\!r$. During the image mapping process, the parameters of the local region size $r$ and the output stride $s$ are critical since they determine whether there is redundancy in the map. Throughout the experimental process, it is necessary to ensure that $r\!\geq\!s$. When $r\!>\!s$, every two adjacent local regions overlap with $\frac{r-s}{r}$, the count map becomes redundant in this situation.
The overlap diminishes only when $r\!=\!s$. In our paper, the parameters of r and s are defined as 64 and 8, respectively. Thus, the resulting count map exhibits redundancy. The normalization step must be performed to eliminate redundancy and ensure that the final sum of normalized count maps accurately reflects the plant count of the image. The base input size of the network is $r\!\times\!r$, which is only tied to the network architecture. And owing to the architecture, the input size of the images can be arbitrary. For instance, the input image  $\mathbf{I}\!\in\!\mathbb{R}^{H\times W\times3}$, CountMamba defines a transformation $f$ as follows: $f(\mathbf{I}): \mathbb{R}^{H\times W\times3}\!\rightarrow\!\mathbb{R}^{\frac{H}{s}\times\frac{W}{s}}$, where $H, W \gg r$.

\begin{table}[t]
\caption{Performance on the MTC dataset. $\dagger$ indicates our re-implementation.}\label{tab:main_results_mtc}
\renewcommand{\arraystretch}{1.15}
\begin{tabular}{|p{3.3cm}|p{1.5cm}<{\centering}|p{1.5cm}<{\centering}|p{1.65cm}<{\centering}|p{1.85cm}<{\centering}|p{1.5cm}<{\centering}|}
\hline
\multicolumn{1}{|c|}{\textbf{Method}} & \textbf{MAE} & \textbf{RMSE} & \textbf{rMAE(\%)} & \textbf{rRMSE(\%)} & $\mathbf{R^2}$ \\ \hline

MCNN \cite{zhang2016single} & 17.9 & 21.9 & 273.4 & 692.5 & 0.33 \\
CSRNet \cite{li2018csrnet} & 6.9 & 11.5 & 77.8 & 190.3 & 0.82 \\
SANet \cite{cao2018scale}& 5.1 & 11.4 & 28.7 & 60.9 & - \\
BCNet \cite{liu2019counting} & 5.2 & 9.2 & 31.8 & 62.5 & 0.88 \\
SFC2Net \cite{liu2020high} & 5.0 & 9.4 & 17.7 & 24.6 & 0.89 \\
TasselNet \cite{lu2017tasselnet} & 6.6 & 9.9 & 44.8 & 89.9 & 0.87 \\
TasselNetV2 \cite{xiong2019tasselnetv2} & 5.4 & 9.2 & 31.9 & 69.5 & 0.89 \\
TasselNetV2+ \cite{xiong2019tasselnetv2} & 5.1 & 9.1 & - & - & 0.89 \\
TasselNetV2+${^\dagger}$ \cite{xiong2019tasselnetv2} & 4.9 & 8.5 & 25.8 & 46.4 & 0.90 \\
STEERER \cite{han2023steerer} & 5.4 & 8.1 & 44.7 & 107.0 & 0.89 \\
CountMamba (Ours) & 4.6 & 7.9& 26.2 & 49.9 & 0.92 \\ \hline

\end{tabular}
\end{table}

\begin{table}[t]
\caption{Performance on the WED dataset. $\dagger$ indicates our re-implementation.}\label{tab:main_results_wed}
\renewcommand{\arraystretch}{1.15}
\begin{tabular}{|p{3.3cm}|p{1.5cm}<{\centering}|p{1.5cm}<{\centering}|p{1.65cm}<{\centering}|p{1.85cm}<{\centering}|p{1.5cm}<{\centering}|}
\hline
\multicolumn{1}{|c|}{\textbf{Method}} & \textbf{MAE} & \textbf{RMSE} & \textbf{rMAE(\%)} & \textbf{rRMSE(\%)} & $\mathbf{R^2}$\\ \hline
MCNN \cite{zhang2016single} & 11.5 & 15.6 & 8.4 & 11.0 & 0.38 \\
CSRNet \cite{li2018csrnet} & 4.2 & 5.2 & 3.2 & 4.1 & 0.94 \\
SANet \cite{cao2018scale}& 4.9 & 6.2 & 3.9 & 5.0 & - \\
BCNet \cite{liu2019counting} & 4.1 &4.9 & 3.1 & 3.8 & 0.94 \\
TasselNet \cite{lu2017tasselnet} & 6.8 & 8.3 & - & 7.1 &0.79  \\
TasselNetV2 \cite{xiong2019tasselnetv2} & 5.3 & 6.8 & 4.1 & 5.3 &0.90  \\
TasselNetV2+ \cite{xiong2019tasselnetv2} & 4.9 & 6.1 & - & 4.6 & 0.91 \\
TasselNetV2+${^\dagger}$ \cite{xiong2019tasselnetv2} & 4.8 & 5.9 & 3.7 & 4.6 & 0.91 \\
SFC2Net \cite{liu2020high} & 4.2 & 5.1 & 3.2 & 4.2 & 0.93 \\
STEERER \cite{han2023steerer} & 5.8 & 6.8 & 4.3 & 5.2 & 0.89 \\
CountMamba (Ours) & 5.3 & 6.5 & 4.0 & 4.9 & 0.89 \\ \hline

\end{tabular}
\end{table}

As a redundant count map $\mathbf{C}_{r}\!\in\!\mathbb{R}^{\frac{H}{s}\times\frac{W}{s}}$ is emerged after the operation $f$, the normalized count map $\mathbf{C}_{n}\!\in\!\mathbb{R}^{H\times W}$ is designed. Subsequently, the image-level count $C$ can be calculated by aggregating $\mathbf{C}_n$, as follows:
\begin{equation}\label{count}
    C=\sum_{x=1}^{W}\sum_{y=1}^{H}\mathbf{C}_{n}(x, y)
\end{equation}
where $\mathbf{C}_{n}(x, y)$ represents each element of $\mathbf{C}_{n}$ indexed by $x$ and $y$.

\subsection{Loss Function}
Following previous works \cite{lu2017tasselnet,xiong2019tasselnetv2}, we optimize our CountMamba with $L_1$ loss for counting values, which can be formulated as:
\begin{equation}
    L=\frac{1}{B}\sum_{i=1}^{B}|C_{i}-{C}_{i}^{*}|
\end{equation}
where $B$ is the batch size, $C_{i}$ and ${C}_{i}^{*}$ are the predicted number of plants and the ground truth in the $i$-th image, respectively.

\section{Experiments}
\subsection{Dataset and Evaluation Metric}

\textbf{MTC dataset} is a maize tassels count dataset first introduced by Liu et al. \cite{lu2017tasselnet}. The maize tassels in the MTC dataset exhibit significant variations in scale and shape. Their original resolution ranges from $3648\times2736$, $4272\times2848$, and $3456\times2304$. For the experiment, 186 images are randomly selected and utilized as the training set, while the remaining 175 images are allocated to the test set. Every maize tassel in the dataset is manually labeled using a bounding box. These bounding-box annotations are then converted into dot annotations by calculating their central coordinates.

\textbf{WED dataset} is a wheat ear dataset first introduced by Madec et al. \cite{madec2019ear}. The images have a resolution of $6000\times4000$ pixels, with the number of ears varying from 80 to 170 in each image. The dataset contains 236 images, with 165 allocated for training and 71 for testing. While bounding box annotations are provided in this dataset, only the center point of each box is utilized to ensure the uniform comparison of different methods for this experiment.

\textbf{SHC dataset} is a sorghum heads counting dataset first introduced by Guo et al. \cite{madec2019ear}. It consists of two subsets with 52 and 40 images, respectively. All the images are labeled with dotted annotations. Our experiments are evaluated on dataset 1, in which 26 images are randomly selected for training and the remaining for testing. We follow the same training configuration used in \cite{xiong2019tasselnetv2}.

To evacuate the counting performance on these above datasets, we adopted the $\text{MAE}$, $\text{RMSE}$, $\text{rMAE}$, $\text{rRMSE}$, and $\text{R}^{2}$ as the basic metrics. These metrics are calculated as follows:

\begin{equation}
    \text{MAE} = \frac{1}{N} \sum_{i=1}^{N}\left | G_{i} - P_{i} \right | 
\end{equation}

\begin{equation}
   \text{RMSE} = \sqrt{\frac{1}{N} \sum_{i=1}^{N} ( G_{i} - P_{i})^2}  
\end{equation}

\begin{equation}
   \text{rMAE} = \frac{1}{N} \sum_{i=1}^{N}\left | \frac{G_{i}-P_{i}}{G_{i}} \right | 
\end{equation}

\begin{equation}
   \text{rRMSE} = \sqrt[]{\frac{1}{N} \sum_{i=1}^{N}( \frac{G_{i}-P_{i}}{G_{i}})^2} 
\end{equation}

\begin{equation}
   \text{R}^{2}=1-\frac{\sum_{i=1}^{N}\left(P_{i}-G_{i}\right)^{2}}{\sum_{i=1}^{N}\left(P_{i}-\bar{G}\right)^{2}}
\end{equation}
where $G_{i}$ is the real number of a plant in the ith image, $P_{i}$ is the prediction number of the plant in the ith image, $N$ is the number of total images, and $\bar{G}$ is the mean real number count.

\subsection{Implementation Details}

All the images in the MTC dataset are resized to 512 $\times$ 512, and crop size is set to 256 $\times$ 256. As for the WED dataset, the down-sampling ratio is set to $\frac{1}{8}$, the crop size is set to 256 $\times$ 256. All the images in the SHC dataset are randomly cropped as 256 $\times$ 1024. The training process employs Adam optimizer with an initial learning rate of 1$\times$$10^{-4}$. The size of image patches is $2\times2$, and the hyperparameter $\beta$ is set to 1. All experiments are carried out with the Pytorch framework and a single NVIDIA GeForce RTX 4090 GPU.

\section{Results and Analysis}

\subsection{Comparison with the State-of-the-Arts}
Tab. \ref{tab:main_results_mtc}, \ref{tab:main_results_wed}, and \ref{tab:main_results_shc} show the quantitative results between CountMamba and state-of-the-art plant counting methods on MTC, WED and SHC dataset, respectively. Thanks to the adaptation to the distribution of different directions, our proposed CountMamba achieves competitive performance on all three benchmark datasets. Notably, CountMamba achieves the best $\text{MAE}$ of 4.6 and $\text{RMSE}$ of 7.9 on the MTC dataset and the best $\text{RMSE}$ of 18.6 on the SHC dataset. 

\begin{table}[t]
\caption{Performance on the SHC dataset. $\dagger$ indicates our re-implementation.}\label{tab:main_results_shc}
\renewcommand{\arraystretch}{1.15}
\begin{tabular}{|p{3.3cm}|p{1.5cm}<{\centering}|p{1.5cm}<{\centering}|p{1.65cm}<{\centering}|p{1.85cm}<{\centering}|p{1.5cm}<{\centering}|}
\hline
\multicolumn{1}{|c|}{\textbf{Method}} & \textbf{MAE} & \textbf{RMSE} & \textbf{rMAE(\%)} & \textbf{rRMSE(\%)} & $\mathbf{R^2}$\\ \hline
TasselNetV2 \cite{xiong2019tasselnetv2} & 18.0 & 21.3 & - & - &0.96  \\
TasselNetV2+ \cite{xiong2019tasselnetv2} & 17.5 & 20.6 & - & - & 0.96 \\
TasselNetV2+${^\dagger}$ \cite{xiong2019tasselnetv2} & 21.4 & 23.7 & 4.4 & 4.9 & 0.94 \\
STEERER \cite{han2023steerer} & 15.6 & 19.2 & 3.2 & 3.9 & 0.95 \\
CountMamba (Ours) & 15.9 & 18.6 & 3.2 & 3.7 & 0.96 \\ \hline

\end{tabular}
\end{table}

\begin{table}[t]
\caption{Ablation study of different scan directions on MTC dataset.}\label{tab:ablation_direction}
\renewcommand{\arraystretch}{1.15}
 \begin{tabular}{|p{1cm}<{\centering}|p{1.1cm}<{\centering}|p{1.1cm}<{\centering}|p{1.1cm}<{\centering}|p{1cm}<{\centering}|p{1.1cm}<{\centering}|p{1.75cm}<{\centering}|p{1.8cm}<{\centering}|p{1cm}<{\centering}|}
\hline
\multicolumn{1}{|c|}{\textbf{HSSM}} & \textbf{VSSM} & \textbf{DSSM} & \textbf{ASSM} & \textbf{MAE} & \textbf{RMSE} & \textbf{rMAE(\%)} & \textbf{rRMSE(\%)} & $\mathbf{R^2}$ \\ \hline

& & & \checkmark & 5.0 & 8.4 & 34.3 & 78.0 & 0.91 \\
& &\checkmark & \checkmark & 4.6 & 7.7 & 30.1 & 59.4 & 0.92 \\
& \checkmark &\checkmark & \checkmark & 4.6 & 7.9 & 27.5 & 55.6 & 0.91 \\
\checkmark &\checkmark &\checkmark& \checkmark & 4.6 & 7.9 & 26.2 & 49.9 & 0.92 \\ \hline
\end{tabular}
\end{table}

\begin{table}[t]
\caption{Ablation study of expert numbers on the MTC dataset.}\label{tab:ablation_expert}
\renewcommand{\arraystretch}{1.15}
\begin{tabular}{|p{1.89cm}<{\centering}|p{1.89cm}<{\centering}|p{1.89cm}<{\centering}|p{1.89cm}<{\centering}|p{1.89cm}<{\centering}|p{1.89cm}<{\centering}|}

\hline
\multicolumn{1}{|c|}{\textbf{\#Experts}} & \textbf{MAE} & \textbf{RMSE} & \textbf{rMAE(\%)} & \textbf{rRMSE(\%)} & $\mathbf{R^2}$ \\ \hline

One & 4.9 & 8.3 & 33.6 & 73.9 & 0.91 \\
Two & 4.7 & 8.0 & 28.7 & 58.1 & 0.91 \\
Four & 4.6 & 7.9 & 26.2 & 49.9 & 0.92 \\ \hline
\end{tabular}
\end{table}

\begin{table}[t]
\caption{Ablation study of Global-Local Adaptive Fusion on MTC dataset.}\label{tab:ablation_fusion}
\resizebox{1.0\linewidth}{!}{
\renewcommand{\arraystretch}{1.15}
\begin{tabular}{|c|c|c|c|c|c|c|c|c|}
\hline
\multicolumn{1}{|c|}{\textbf{Adaptive Fusion}} & \textbf{CNN Branch} & \textbf{MAE} & \textbf{RMSE} & \textbf{rMAE(\%)} & \textbf{rRMSE(\%)} & $\mathbf{R^2}$ \\ \hline

&  & 6.5 & 10.9 & 45.4 & 91.1 & 0.85 \\
& \checkmark & 4.9 & 8.2 & 33.7 & 69.4 & 0.91 \\
\checkmark& \checkmark & 4.6 & 7.9 & 26.2 & 49.9 & 0.92 \\\hline
\end{tabular}
}
\end{table}

\begin{figure}[t]
    \centering
    \includegraphics[width=0.99\linewidth]{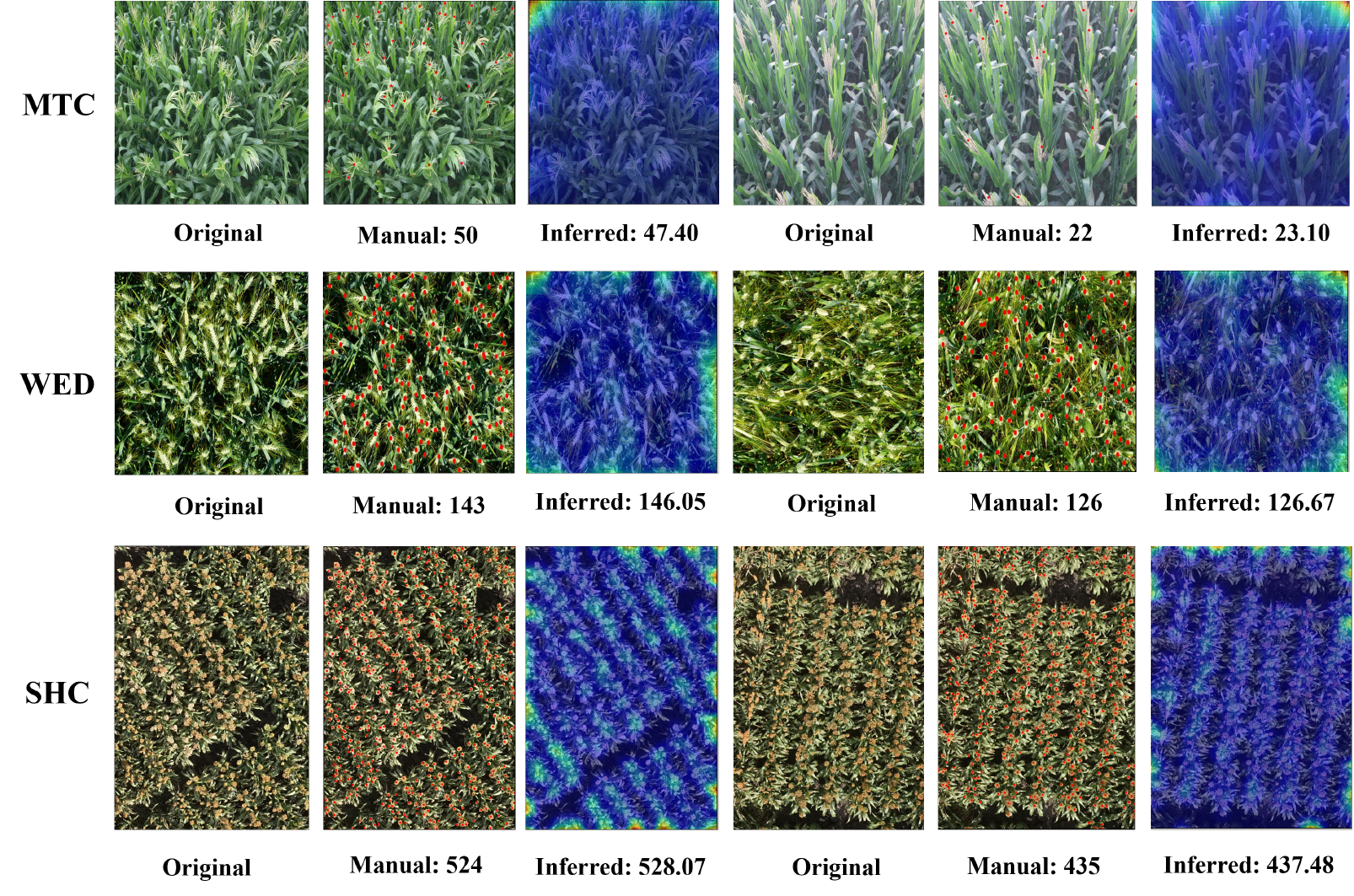}
    \caption{Qualitative results on MTC, WED, and SHC datasets. \textit{Manual} indicates the ground-truth and \textit{Inferred} the predicted count. Red points are manual annotations.}
    \label{fig:visualization}
\end{figure}

\subsection{Ablation Study}
\textbf{Effects of different scan directions.} To allow Mamba to count plants distributed in various directions, we build four SSMs inspired by \cite{zhao2024rs}, which uses scans in eight directions to generate scanned sequences. Here, we ablate different scan directions to study the effects, and the results are shown in \ref{tab:ablation_direction}. We are adding anti-diagonal, diagonal, vertical, and horizontal scanning in succession, which results in better results, demonstrating that counting in different order benefits the plant distribution in various directions.

\textbf{Effects of expert numbers.} We compare different numbers of counting experts in Tab. \ref{tab:ablation_expert}, including 1) One: one State-Space Block that scans in the horizontal, vertical, diagonal, and anti-diagonal directions simultaneously; 2) Two: one State-Space Block that scans in the horizontal and vertical directions and the other in the diagonal and anti-diagonal directions; 3) Four: using HSSB, VSSB, DSSB and ASSB. The results show that assigning feature extraction in the horizontal, vertical, diagonal, and anti-diagonal directions to four experts achieves the best counting results.

\textbf{Effects of Global-Local Adaptive Fusion.}
We ablate two key designs in Global-Local Adaptive Fusion. The results, presented in \ref{tab:ablation_fusion}, indicate that 1) Introducing a CNN branch to complement the global relationships with local information can effectively enhance the fine-grained local details. 2) Without using sample-wise adaptive fusion, i.e., directly using the average of HSSM, VSSM, DSSM, and ASSM for counting, can only obtain sub-optimal results. Using sample-wise adaptive fusion can flexibly adapt to plants with different spatial distributions. 

\subsection{Visualization}
Qualitative results of our proposed CountMamba on MTC, WED, and SHC datasets are shown in Fig. \ref{fig:visualization}. We observe that: 1) CountMamba obtains strong responses on plant regions and weak responses on non-plant regions in the counting map and thus can infer accurate counting results. 2) CountMamba is robust to the direction of plant distribution and consistently effectively infers numbers from tens to hundreds.

\section{Conclusion and Future Work}
In this paper, we propose CountMamba to explore the power of the recent advanced state space model for plant counting. Our method leverages a Multi-directional State-Space Group to scan image patches in multiple orders, adapting to any distribution of plants. Moreover, Global-Local Adaptive Fusion is utilized to adaptively aggregate the global features extracted from various directions and local features extracted by the CNN branch in a sample-wise manner. Extensive experiments on multiple benchmark datasets demonstrate our CountMamba serves as a simple but effective state-space model for plant counting. Future work will explore the scalability of our approach to more complex and diverse plant scenarios, as well as the potential integration of fine-grained feature extraction strategies.

\vspace{\baselineskip}

\noindent\textbf{Acknowledgments.} This work was supported by the grants from the National Natural Science Foundation of China (U22B2048, 61925201, 62132001, 62373043).

\bibliography{ref}

\end{document}